\documentclass[conference]{IEEEtran}
\IEEEoverridecommandlockouts
\usepackage{hyperref}
\usepackage{amsmath,amssymb,amsfonts}
\usepackage{multirow}
\usepackage{subfig}
\usepackage{algorithmic}
\usepackage{graphicx}
\usepackage{textcomp}
\usepackage{xcolor}
\def\BibTeX{{\rm B\kern-.05em{\sc i\kern-.025em b}\kern-.08em
    T\kern-.1667em\lower.7ex\hbox{E}\kern-.125emX}}

\begin{document}
\title{Deep Fake Detection : Survey of Facial Manipulation Detection Solutions}
\author{
\IEEEauthorblockN{\footnotesize Samay Pashine}
\IEEEauthorblockA{\footnotesize \textit{Dept. of Computer Science} \\
\textit{{\footnotesize AITR}}\\
\footnotesize Indore, India \\
\footnotesize samaypashine7@gmail.com
\and
\IEEEauthorblockN{\footnotesize Sagar Mandiya}
\IEEEauthorblockA{\footnotesize \textit{Dept. of Computer Science} \\
\textit{{\footnotesize AITR}}\\
\footnotesize Indore, India \\
\footnotesize sagar.mandiya15@gmail.com}
\and
\centering \IEEEauthorblockN{\footnotesize Praveen Gupta}
\IEEEauthorblockA{\footnotesize \textit{Dept. of Computer Science} \\
\textit{{\footnotesize AITR}}\\
\footnotesize Indore, India \\
\footnotesize praveengupta090@gmail.com}
\and
\IEEEauthorblockN{{\footnotesize Prof. Rashid Sheikh}}
\IEEEauthorblockA{\footnotesize \textit{Dept. of Computer Science} \\
\textit{{{\footnotesize AITR}}}\\
\footnotesize Indore, India \\
\footnotesize rashidsheikh@acropolis.in} \\
\and
\IEEEauthorblockN{\footnotesize}
\IEEEauthorblockA{\footnotesize} \\
\textit{{\footnotesize}}\\
\footnotesize \\
\footnotesize}
}

\maketitle

\begin{abstract}
Deep Learning as a field has been successfully used to solve a plethora of complex problems, the likes of which we couldn't have imagined a few decades back. But as many benefits as it brings, there are still ways in which it can be used to bring harm to our society. Deep fakes have been proven to be one such problem, and now more than ever, when any individual can create a fake image or video simply using an application on the smartphone, there need to be some countermeasures, with which we can detect if the image or video is a fake or real and dispose of the problem threatening the trustworthiness of online information. Although the Deep fakes created by neural networks, may seem to be as real as a real image or video, it still leaves behind spatial and temporal traces or signatures after moderation, these signatures while being invisible to a human eye can be detected with the help of a neural network trained to specialize in Deep fake detection. \\

In this paper, we analyze several such states of the art neural networks (MesoNet, ResNet-50, VGG-19, and Xception Net) and compare them against each other, to find an optimal solution for various scenarios like real-time deep fake detection to be deployed in online social media platforms where the classification should be made as fast as possible or for a small news agency where the classification need not be in real-time but requires utmost accuracy. \\
~ \\
Github link: \href{https://github.com/sagarmandiya/DeepFake-Detection}{[github.com/sagarmandiya/DeepFake-Detection]} \\

Keywords: Deep Learning, Neural Networks, Deep Fakes, MesoNet, ResNet, VGG, Xception\\
\end{abstract}

\section{\textbf{INTRODUCTION}}
Forgery and manipulation of multimedia like images and videos including facial information generated by digital manipulation, in particular with DeepFake methods, have become a great public concern recently \cite{citrond}, \cite{rcellanjones} especially for public figures. The famous term “DeepFake” is referred to a deep learning-based technique able to create fake videos by manipulating features or swapping the face of a person by the face of another person. This term originated after a Reddit user named “deepfakes” claimed in late 2017 to have developed an algorithm that helped him to transpose celebrity faces into adult videos \cite{bbcbitesize}. Additionally, to fake pornography, some of the more harmful usages of such fake content include fake news, hoaxes, financial fraud, and defamation of the victim. Resulting in revitalizing general media forensics which is now dedicated to advance in detecting facial manipulation in image and video \cite{swaminathan2008digital} \cite{korus2017digital} \cite{rossler2019faceforensics++}.\\

The efforts in fake face detection are built on the foundation of the past research in biometric anti-spoofing and modern supervised deep learning \cite{neves2020ganprintr} \cite{dang2020detection}. The growing interest in manipulation detection is demonstrated through the increasing number of workshops in various top conferences. International projects such as MediFor funded by the DARPA, and competitions such as the Media Forensics Challenge and the Deepfake Detection Challenge launched by the National Institute of Standards and Technology (NIST) and Facebook, respectively. In the old days, the number and realism of the manipulations have been limited by the lack of advanced tools, domain expertise, and the complex and time-consuming process. For example, the early work in this domain \cite{bregler1997video} was able to modify the motion of the lip using a different audio track, by making connections between the soundtrack and the shape of the subject’s face. However, many things have evolved now since those experiments. Nowadays, it is becoming really easy to synthesize/generate non-existent faces or manipulate an existing face in an image/video, All of this is possible because of accessibility to large-scale public data, and the advancement in deep learning techniques that eliminate many manual steps such as Autoencoders (AE) and Generative Adversarial Networks (GAN) \cite{kingma2013auto}, \cite{goodfellow2014generative}. As a result, Much public software and mobile application (e.g FaceApp, etc) have been released giving access to everyone to create fake images and videos, without any experience in this domain. Therefore, to counter those advanced and realistic manipulated content, large efforts are being carried out by the research community to design improved methods for face manipulation detection. \\

Over the past couple of years, huge steps forward in the field of automatic video editing techniques have been made and great interest has been shown towards methods for facial manipulation. For Instance, it is nowadays possible to easily perform facial reenactment, i.e. transferring the facial expressions between people. This enables to change of the identity of a speaker with almost no effort. Advancement in these Systems and tools for facial manipulations enables even users without any previous experience in digital arts to use them. Indeed, code and libraries that work in an almost automatic fashion are more and more often open sources. On one hand, this technological advancement opens the door to uncharted territories. And On the other hand, people are using these gifts in the worst possible ways for their reasons. \\

In this paper, We consider MesoNet, ResNet-50, VGG-19, Xception and comparing their characteristics to know which of these networks is the most efficient and accurate one on basis of different parameters like operation time, accuracy rate, loss rate, and ability to perform on random data. Training and Evaluation are performed on three datasets: Celeb-DF and Celeb-DF-v2, which has been proposed as a public benchmark; DFDC, which has been released as part of the DFDC Kaggle competition. Results show that the attention-based neural network modification helps the system in outperforming the baseline reported in the domain on all three datasets. Our paper makes contributions by Comparing the state-of-the-art neural networks like MesoNet, VGG-19, ResNet-50, and Xception performances in this domain and drawing the conclusion from the results to advance in media manipulation detection. Detailed evaluation of complex forgery detectors in various scenarios.\\

\section{\textbf{PROBLEM FORMULATION}}
The recent improvement in the field of Deep Learning has produced some state-of-the-art neural network architectures like xception network(sometimes also referred to as extreme inception), Senet, and others, which in turn has lead to astonishing developments in the field of machine learning and computer vision. Although the benefits of such an invention far outweigh the cons, there still exists some which if not treated in time can lead to major disarray in society as we know it. One such con is the creation of deep fakes, deep fakes are computer-generated fake images and videos, which as of today floods one of the major sources of information i.e., the Internet. If not treated it can lead to some major problems, of which privacy violation, public defamation are few. A recent Forbes article \cite{robtoews} claims that "Deepfakes Are Going To Wreak Havoc On Society. We Are Not Prepared.", Currently, the predominant use of deepfakes is for pornography. In June 2020, research \cite{adamsmith} indicated that 96 percent of all deepfakes online are for pornographic context, and nearly 100 percent of those cases are of women and that many actresses like Kristen bell have already suffered from it.\\

All of these incidents beg to raise the question that, what have we done to stop this and the answer lies in deep fake detection. In layman terms deep fake detection is done by neural networks specializing in detecting deepfakes, i.e., With deep fake detection we can detect if a photo or video is fake or real, and therefore it must remain an active topic for research so that we can filter out the fake content from the internet, and once again make it reliable. Many of the tech giants like Facebook has also taken initiatives to try and stop this misuse of neural networks which otherwise is a wonderful technology. \\

\section{\textbf{RELATED WORK}}
In the last couple of years, several techniques for facial manipulation in media like images, video, etc have been successfully developed and are available to the public (i.e., FaceSwap, Face2Face, deepfake, etc.). This enables anyone to easily edit faces in video sequences with incredibly realistic results and very little effort. Moreover, the free access to large-scale public databases, together with the fast progress of deep learning techniques, in particular Generative Adversarial Networks, have led to the generation of very realistic fake content with its corresponding implications towards society in this era of fake news. Likewise, deepfake detection is also an important application of deep learning and machine learning which helps detect forgeries in media like images, and videos and a wide range of research has already been done that encompasses a comprehensive study and implementation of various popular algorithms. \cite{bonettini2021video} where They tackle the problem of face manipulation detection in video sequences targeting modern facial manipulation techniques. In particular, the ensembling of different trained Convolutional Neural Network (CNN) models. In the proposed solution, different models are obtained starting from a base network making use of two different concepts, attention layers, and siamese training. They showed the community that combining these networks leads to promising face manipulation detection results on two publicly available datasets with more than 119000 videos. In \cite{tolosana2020deepfakes} the authors survey the other popular techniques for manipulating face images including DeepFake methods, and methods to detect such manipulations. In particular, they reviewed four types of facial manipulation, entire face synthesis, identity swap (DeepFakes), attribute manipulation, and expression swap. For each of them, they provided details regarding manipulation techniques on existing open-source databases, including a summary of results from those evaluations. \\

\section{\textbf{METHODOLOGY}}
The comparison of the neural networks (MesoNet, ResNet-50, VGG-19, and Xception) is based on the characteristic chart of each network on common grounds like dataset, the number of epochs, complexity of the network, accuracy of each network, specification of the device (Ubuntu 20.04 LTS, 8 GB RAM, intel core i7 8th gen processor, NVIDIA GTX 1050Ti GPU) used to execute the program and runtime of the algorithm, under ideal condition. \\

\begin{flushleft}
\textbf{\textit{A. 		DATASET}}
\end{flushleft}
Deep Fake Detection is an expansive research area that already contains detailed ways of implementation which include major learning datasets, popular algorithms, features scaling, and feature extraction methods. Celeb-DF, Celeb-DF-v2, and DFDC datasets are datasets containing the real and manipulated videos of common people and public figures. Due to hardware limitations, we had to take only a small part of the datasets mentioned above. Celeb-DF \& Celeb-DF-v2 are high-quality, large-scale challenging datasets for deepfake forensics. They contain DeepFake videos of celebrities generated using an improved synthesis process. The DFDC dataset was created by the companies to solve the deepfake detection problem and it is by far the largest currently and publicly available face swap video dataset, with over 100,000 total clips sourced from 3,426 paid actors, produced with several Deepfake, GAN-based, and non-learned methods. Celeb-DF contains a total of 1,171 videos out of which 376 are real and 795 are fake. whereas Celeb-DF-v2, DFDC contains 2,172 videos(890 real \& 1,282 fake), and 910 videos(362 real \& 548 fake) respectively. \\
\begin{flushleft}
\includegraphics[scale=0.30]{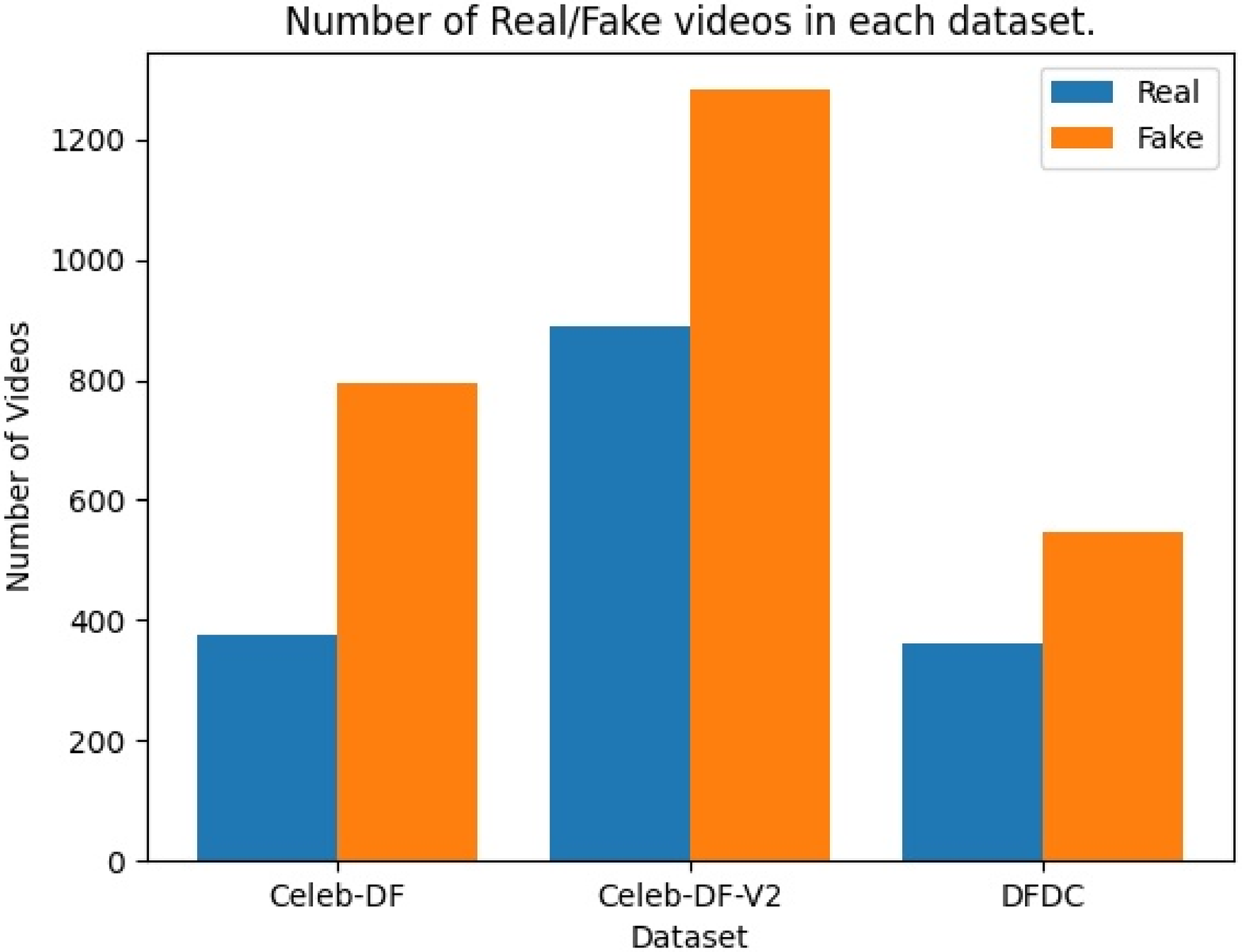}
\end{flushleft}
\footnotesize Figure 1. Category wise number of videos in each dataset that we have used. \\

\begin{center}
\includegraphics[scale=0.30]{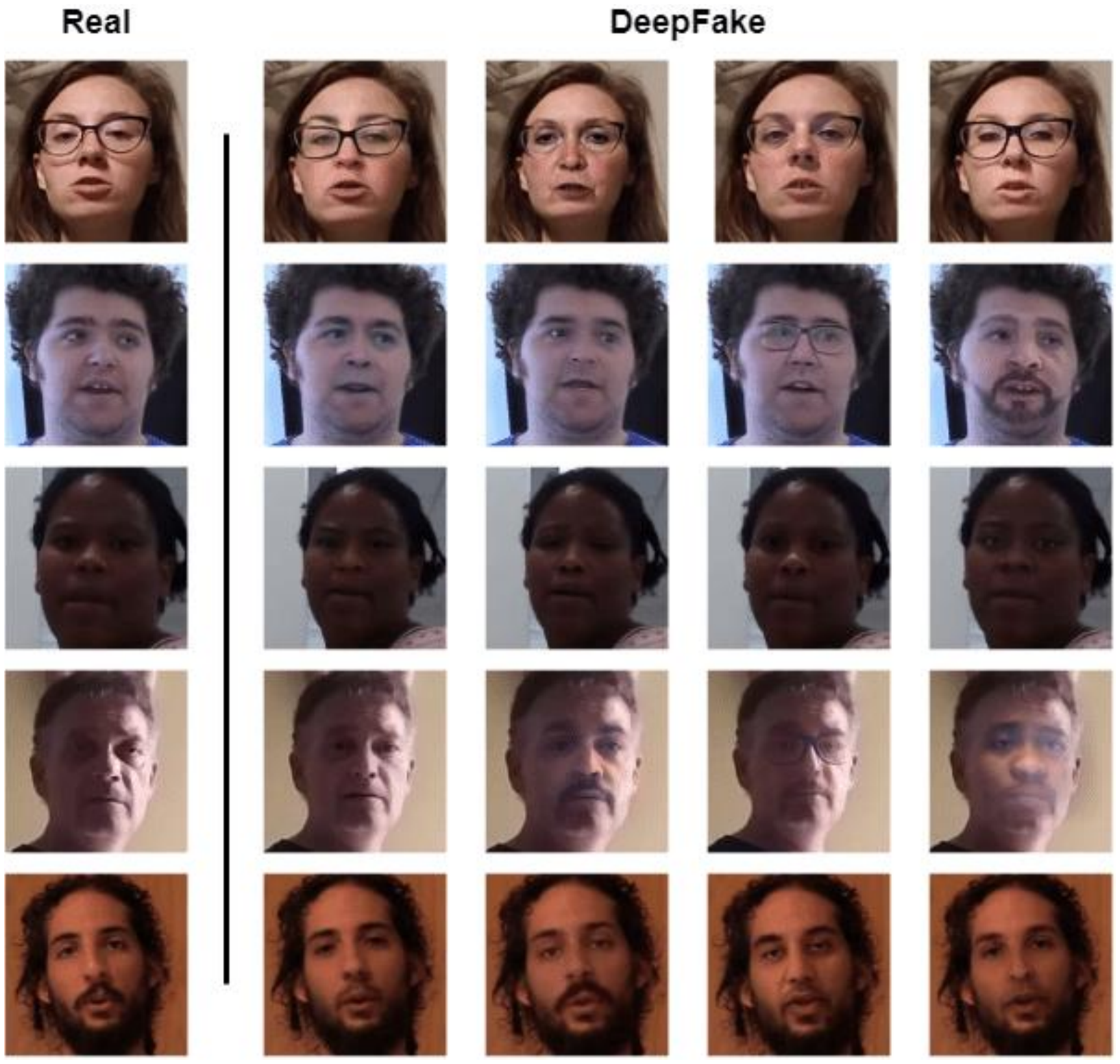}
\footnotesize Figure 2. Some random snapshots of videos from each datasets (Celeb-DF, Celeb-DF-v2, and DFDC). \\
\end{center}

\begin{flushleft}
\textbf{\textit{B.		MESO NETWORK (MesoNet)}}
\end{flushleft}
This network is a derivation from well-performing networks for classification that alternate layers of convolutions, pooling, and a dense network for classification. This neural network comprises a sequence of four convolution layers and pooling and is followed by a fully connected dense layer with one hidden layer in between. The convolutional layers use ReLU as its activation functions that introduce non-linearities and Batch Normalization \cite{ioffe2015batch} to regularize their output which prevent the vanishing gradient problem, and the fully-connected layers use Dropout \cite{srivastava2014dropout} to regularize which improve its robustness and taking generalization on another level \cite{afchar2018mesonet}.

\begin{center}
\includegraphics[scale=0.37]{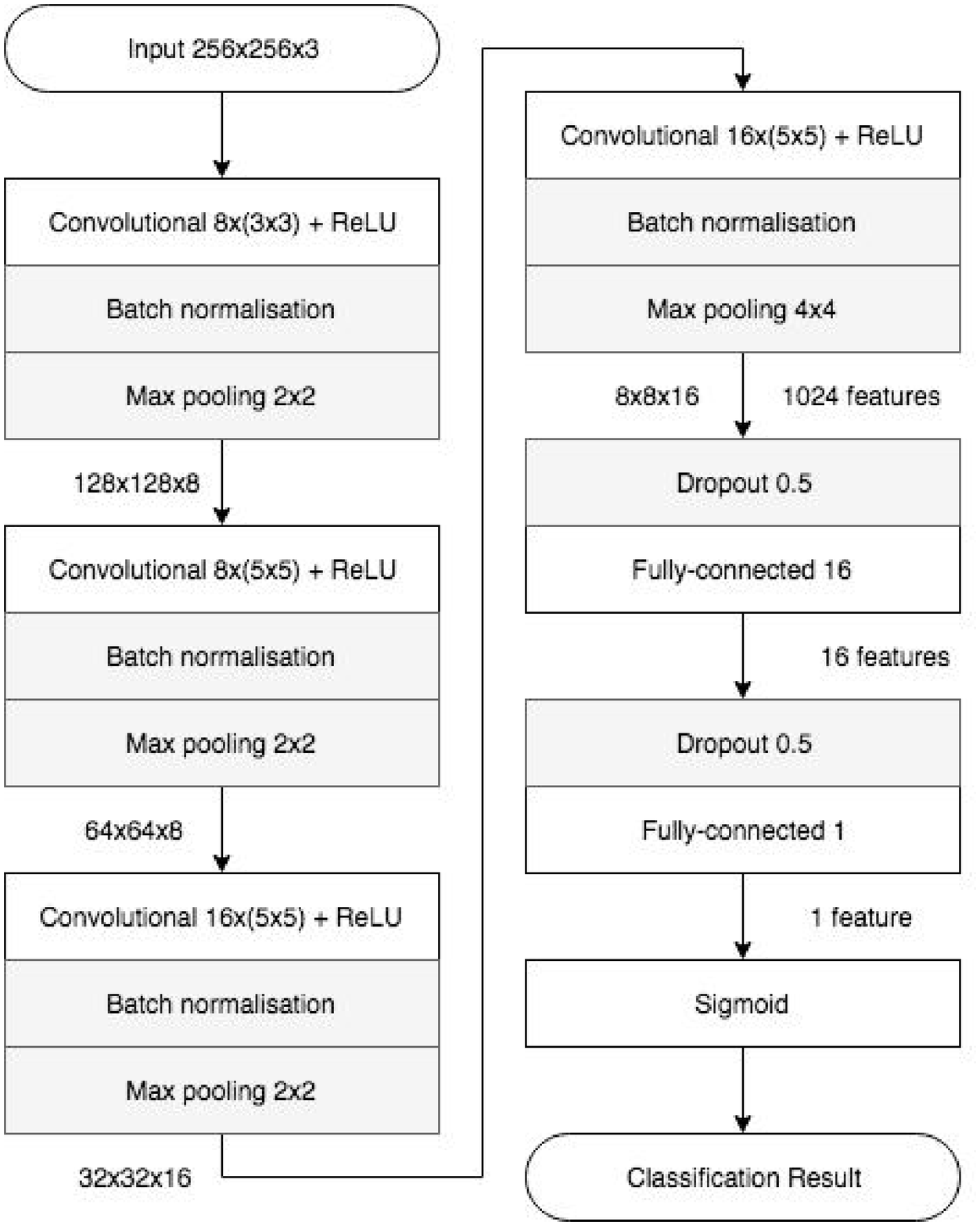}
\end{center}
\footnotesize Figure 3. The network architecture of Meso-4. Layers and parameters. \\

\begin{flushleft}
\begin{normalsize}
\textbf{\textit{C.		RESIDUAL NETWORK (ResNet)}}
\end{normalsize}
\end{flushleft}
Residual Network a.k.a ResNet50 is a variant of the ResNet model which consists of 48 Convolution layers along with 1 MaxPool and 1 Average Pool layer. It is capable of 3.8 billion Floating-point operations. Out of all other variants of residual network with different capabilities, this one widely used ResNet model and we have shown ResNet50 architecture in detail in Figure 4. Because of this framework, it is possible to train ultra DNN (deep neural networks) i.e. Now, the network can contain thousands of layers and still achieve great performance. The ResNets were initially applied to the image recognition task but as is mentioned in the paper that the framework can be used for non-computer vision tasks also to achieve better accuracy. Many people argued that simply stacking more layers also gives us better accuracy why was there a need for Residual learning for training ultra-deep neural networks but stacking more layer arises a serious problem of vanishing/exploding gradients, that is why ResNet is used in this paper so that we can assess it's effectiveness in deepfake detection problem \cite{opengenus1}.

\begin{center}
\includegraphics[scale=0.31]{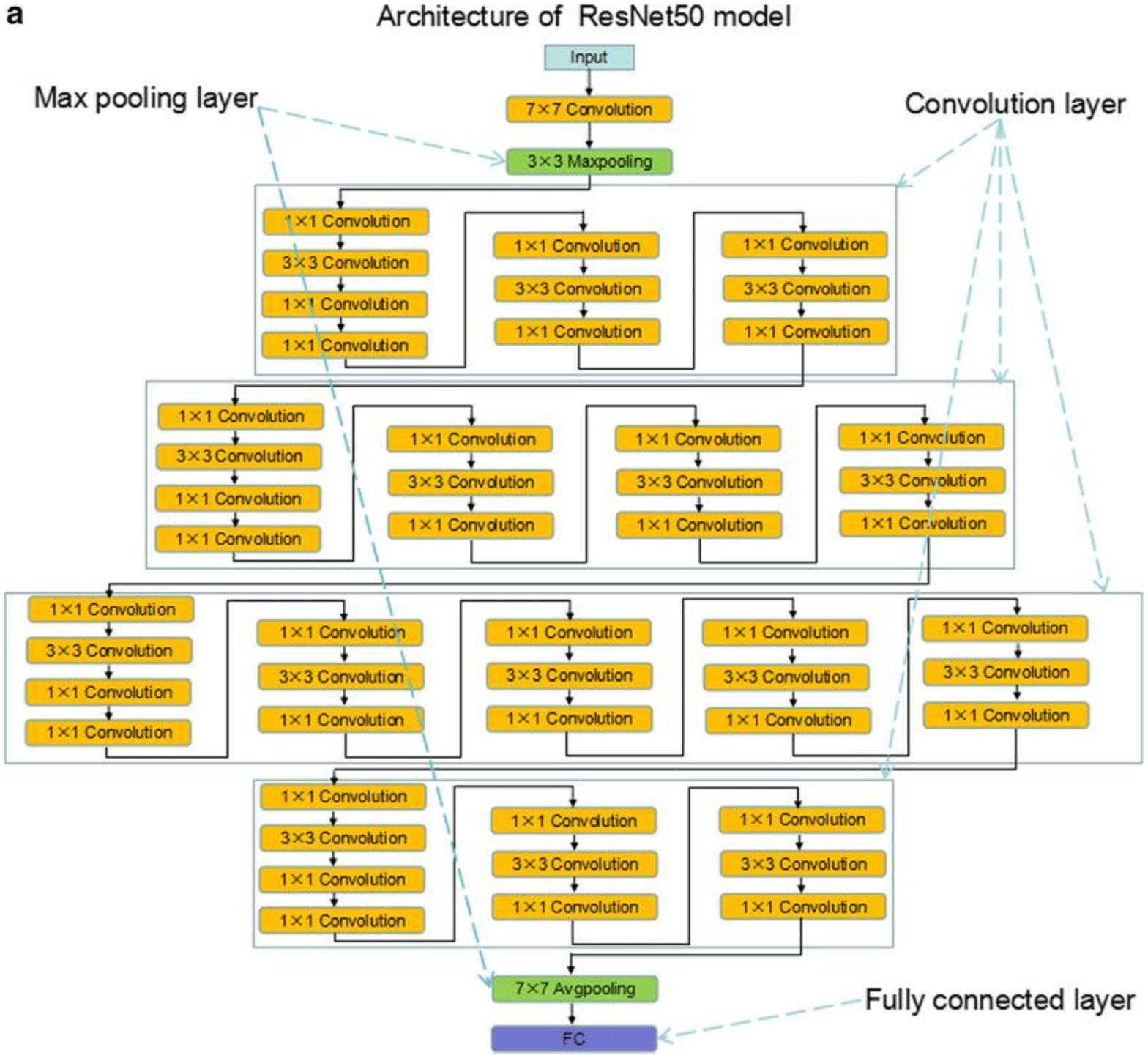}
\end{center}
\footnotesize Figure 4. The architecture of the ResNet-50 with variable specification of the network.\\

\begin{flushleft}
\begin{normalsize}
\textbf{\textit{D.		VISUAL GEOMETRY GROUP NETWORK (VGG-19)}}
\end{normalsize}
\end{flushleft}
Visual Geometry group network a.k.a VGG-19 is a variant of the VGG model which consists of 19 layers that include 16 convolution layers, 3 fully connected layers, 5 MaxPool layers, and 1 SoftMax layer. There are other variants of VGG like VGG-11, VGG-16, etc. VGG-19 has 19.6 billion Floating Operations. VGG is a deep CNN used to classify images \cite{opengenus2}. \\

\begin{center}
\includegraphics[scale=0.29]{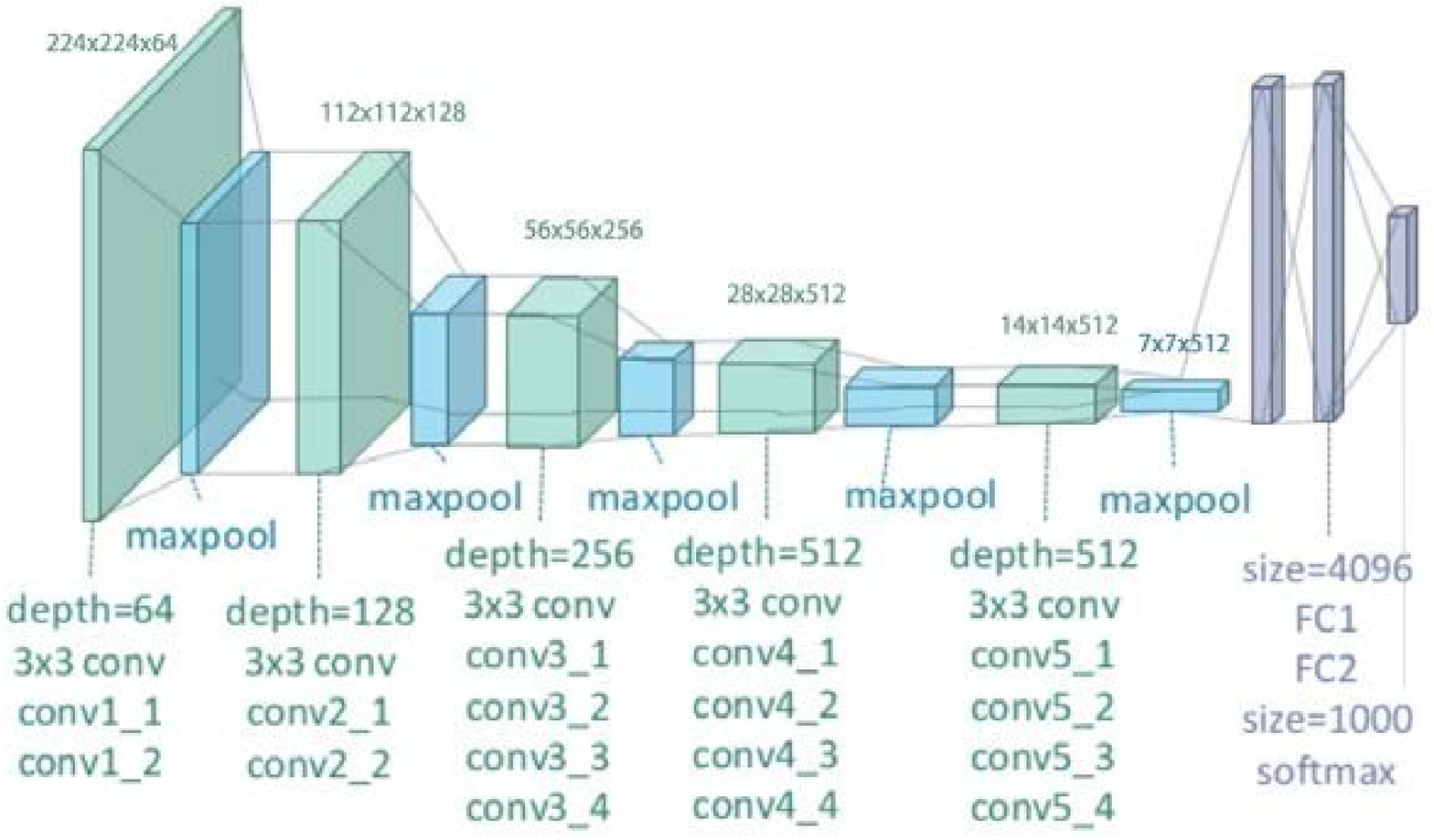}
\end{center}
\footnotesize Figure 5. The architectural design of VGG-19 Network. \\
~ \\

\begin{flushleft}
\begin{normalsize}
\textbf{\textit{E.		XCEPTION NETWORK}}
\end{normalsize}
\end{flushleft}
Xception neural network was created by Google. It stands for Extreme Inception. It consists of a modified depth-wise separable convolution, it has shown even better results than Inception-v3. The original depthwise separable convolution is the depthwise convolution followed by a pointwise convolution but In Xception, modified depthwise separable convolution is the pointwise convolution followed by a depthwise convolution. This modification is motivated by the inception module in Inception-v3. The 14 modules are grouped into three groups viz. the entry flow, the middle flow, and the exit flow. And each of the groups has four, eight, and two modules respectively. The final group, i.e the exit flow, can optionally have fully connected layers at the end. This modification is the reason for the order of operation \& the presence/absence of non-linearity. Due to this modified depthwise separable convolution, there is NO intermediate ReLU non-linearity. Moreover, Xception without any intermediate activation has the highest accuracy. \\

\begin{center}
\includegraphics[scale=0.30]{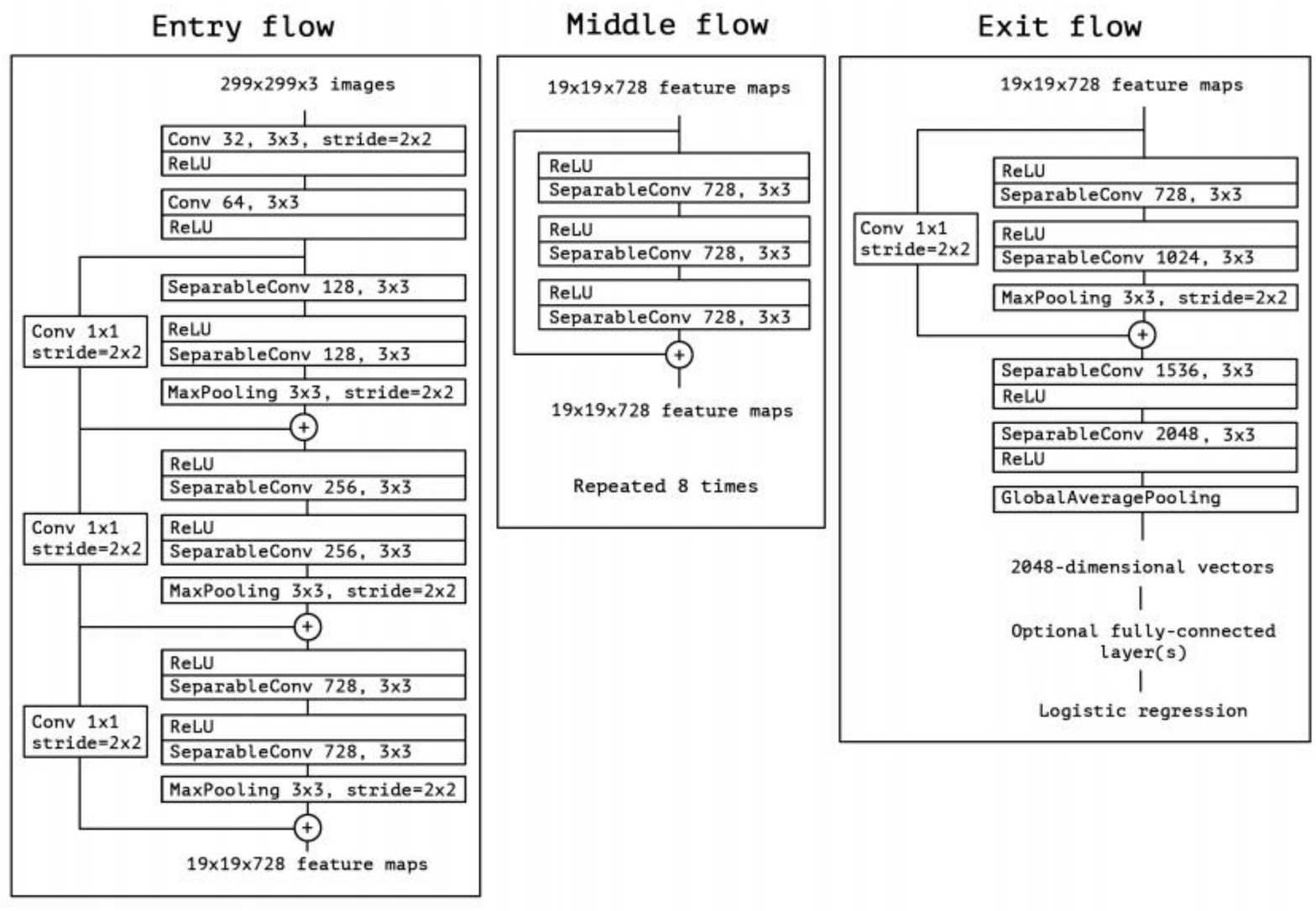}
\end{center}
\footnotesize Figure 6. The architectural design of Xception Network.

\begin{flushleft}
\begin{normalsize}
\textbf{\textit{F.		OPTIMIZATION}}
\end{normalsize}
\end{flushleft}
TensorRT is an SDK for deep learning which provides significantly low inference time, developed by NVIDIA. It contains an inference optimizer and a runtime that is capable of delivers significantly low latency and high throughput for deep learning inference applications. TensorRT-based applications are capable of performing up to a whopping 40 times faster than CPU-only platforms during inference. With TensorRT, you can optimize neural network models trained in all major frameworks, calibrate for lower precision with high accuracy, and deploy to hyper-scale data centers. TensorRT is built on CUDA®, NVIDIA’s parallel programming model which enables the model to efficiently utilize GPU resources, while also enables you to optimize inference leveraging libraries and development tools for artificial intelligence-related tasks. TensorRT provides INT8 and FP16 optimizations for production deployments of deep learning inference applications such as video streaming, speech recognition, recommendation, fraud detection, and natural language processing, to provide the models in class floating point precision. TensorRT achieves this by reducing precision inference significantly which in turn reduces application latency, which is a requirement for many real-time services, as well as autonomous and embedded applications\cite{tensorrt}.

\begin{flushleft}
\begin{normalsize}
\textbf{\textit{G.		VISUALIZATION}}
\end{normalsize}
\end{flushleft}
In this research, we have used multiple datasets (i.e. Celeb-DF, Celeb-DF-v2, and a part of DFDC dataset due to hardware limitations) to compare different neural networks (i.e. MesoNet, ResNet-50, VGG-19, and Xception) based on training \& testing accuracy, training \& testing loss, training time, inference time on CPU, GPU \& after TRT optimization. To visualize the information obtained by the detailed analysis of algorithms we have used Line graphs and Tabular format charts using module matplotlib, which gives us the most precise visuals of the advances of the algorithms in classifying. The graphs are given at each vital part of the programs to give visuals of each part to bolster the outcome. \\

\section{\textbf{IMPLEMENTATION}}
To compare the networks based on working accuracy rate, loss, training time, complexity, and inference time, we have used four different classifiers: \\
\begin{itemize}
	\item MesoNet Classifier 
	\item ResNet-50 : Residual Neural Network
	\item VGG-19 : Visual Geometry Group Network
	\item Xception
\end{itemize}

After training the neural networks, we have optimized the models using TensorRT, to get the minimum inference time and maximum accuracy. We have encapsulated every information in Table 1. \\

\begin{table*}
\centering
\caption{Comparison Analysis of Different network.}
\begin{tabular}{|c|c|c|c|c|c|c|c|} 
\hline
\multirow{2}{*}{Network Name} & \multicolumn{2}{c|}{Training} & \multicolumn{2}{c|}{Testing} & \multicolumn{3}{c|}{Inference Time}   \\ 
\cline{2-8}
                              & ~~~ Accuracy ~~~ & ~~~ Loss ~~~   & ~~~ Accuracy ~~~  & ~~~ Loss ~~~   & ~~~ CPU ~~~  & ~~~ GPU ~~~  & ~~~ TRT Op ~~~   \\ 
\hline
MesoNet                       & 73.189\%           & 25.83       & 72.39\%          & 23.92       & 194 ms       & 180.7 ms    & 64.6 ms             \\ 
\hline
ResNet-50                     & 75.26\%            & 6.55        & 74.12\%          & 15.05       & 1978 ms     & 1142.2 ms   & 789 ms             \\ 
\hline
VGG - 19                      & 74.92\%           & 1.06        & 73.28\%          & 3.39        & 302.2 ms    & 254.3 ms    & 113.9 ms             \\ 
\hline
Xception                      & 77.83\%           & 11.69       & 75.99\%          & 16.11       & 1080 ms     & 1002.1 ms   & 976.2 ms             \\
\hline
\end{tabular}
\\
\end{table*}

We have discussed in detail the implementation of each algorithm explicitly below to create a flow of this analysis for a fluent and accurate comparison.\\

\begin{flushleft}
\textbf{\textit{I. 		DATASET HANDLING \& PRE-PROCESSING}}
\end{flushleft}
The datasets we used in this paper (i.e. Celeb-DF, Celeb-DF-v2, and DFDC) are quite large and due to hardware limitations, we were unable to utilize the complete dataset. So, we took small chunks of the datasets. Now the challenge is to train the neural network on these video datasets. Now, we converted the videos into face images (we used the dlib library to extract images from frames). Overall, We have 51,036 images divided into two categories: Real (19,536 images) and Fake (31,500 images). Since we cannot store all this data for training into the memory, we used the ImageDataGenerator by TensorFlow to create batches of our dataset while training the network. Pre-processing is a crucial step in machine learning which focuses on improving the input data by reducing unwanted impurities and redundancy. To simplify and break down the input data we reshaped all the images present in the dataset in 2-dimensional images i.e (128,128,1). Each pixel value of the images lies between 0 to 255 so, we Normalized these pixel values by dividing them by 255.0 so that the input features will range between 0.0 to 1.0. \\

\begin{flushleft}
\textbf{\textit{II. 	MESO NETWORK}}
\end{flushleft}
The MesoNet-4 used in this paper is a shallow convolutional neural network that was made for the sole purpose of detecting video forgery. In \cite{afchar2018mesonet}, Meso-4 and MesoInception-4 are classes capable of performing binary classification on a dataset. In this paper, we have used MesoNet-4 for the classification of deepfakes datasets. Various libraries and sub-modules of libraries like TensorFlow, TensorFlow.Keras.preprocessing, and matplotlib have been used for the implementation purpose. Firstly, we will download the datasets, followed by loading them using TensorFlow ImageDataGenerator and pre-processing the images while loading them in the network in batches to reduce the memory usage. After this, plotting of some samples of the dataset followed by normalization and scaling of features have been done. Finally, we have created our experimental model. \\

\begin{flushleft}
\textbf{\textit{III. 	RESIDUAL NETWORK - 50}}
\end{flushleft}
The implementation of deepfake detection by ResNet-50 is done with the help of the TensorFlow module to create an MLP model of Sequential class and add the respective inbuilt model of resnet in TensorFlow to take an image of 128x128 pixel size as input. After creating a sequential model, we added a Global average pooling layer followed by a Dense layer. Once you have the training and test data, you can follow these steps to train a neural network in Tensorflow. We used a neural network with 50 hidden layers with multiple max-pooling layers and an output layer with 1 unit (i.e. total number of labels). The number of units in the hidden layers is standard. The input to the network is the 16,384-dimensional array converted from the 128×128 image. We used the Sequential model for building the network. In the Sequential model, we can just stack up layers by adding the desired layer one by one. We used the Dense layer, also called a fully connected layer. Apart from the Dense layer, we added the sigmoid activation function which is a common preference for the binary classification model.

\begin{flushleft}
\textbf{\textit{IV. 	VISUAL GEOMETRY GROUP NETWORK - 19 }}
\end{flushleft}
The model implementation is done using Tensorflow as well. From it, we have used a Sequential class which allowed us to create a model layer-by-layer. The dimension of the input image is set to 128(Height), 128(Width), 3(Number of channels). Next, we added the standard vgg-19 model to this sequential model. The VGG-19 model consists of 19 layers with multiple pooling layers followed by 2 fully connected layers. The pooling layer \cite{machinelearningmastery} is used which reduces the dimensionality of the image and computation in the network. We have employed MAX-pooling which keeps only the maximum value from a pool. The convolution layer uses a matrix to convolve around the input data across its height and width and extract features from it. This matrix is called a Filter or Kernel. The values in the filter matrix are weights. We have used the standard filter of VGG-19. Stride determines the number of pixels shifts. Convolution of filter over the input data gives us activation maps whose dimension is given by the formula: ((N + 2P - F)/S) + 1 where N= dimension of input image, P= padding, F= filter dimension and S=stride. This model returns probability distribution over all the classes. The class with the maximum probability is the output.

\begin{flushleft}
\textbf{\textit{V. 	XCEPTION NETWORK}}
\end{flushleft}
The xception network consists of 36 convolutional layers and its implementation is done using Tensorflow. From Tensorflow, we have used a Sequential class, the dimension of the input image is set to 128(Height), 128(Width), 3(Number of channels). Next, we load the inbuilt standard model of xception. The depthwise separable convolution layer is what powers the Xception. And it heavily uses that in its architecture. This type of convolution is similar to the extreme version of the Inception block. But differs slightly in its working. The effect of having activation on both the depthwise and pointwise steps in the DSC (i.e Deep Separate Convolution). And has observed that learning is faster when there’s no intermediate activation. For this network, we have followed the standard practice \& configuration for training the model.

\section{\textbf{RESULT}}
\begin{small}
After implementing and comparing all the four networks that are MesoNet, ResNet-50, VGG-19, and Xception we have compared their accuracy rate, loss rate, training time, and inference time on both CPU and GPU. Moreover, We also applied TRT optimization on the network models and showed the difference in the performance with the help of experimental graphs for perspicuous understanding. We have taken into account the Training and Testing Accuracy of all the models stated above. Generally, the running time of an algorithm depends on the number of operations it has performed. So, we have trained our large deep learning models like Xception, ResNet, VGG up to 10 epochs (Due to hardware limits), and MesoNet models up to 20 epochs.

\begin{figure}
\centering
\subfloat{{\includegraphics[scale=0.25]{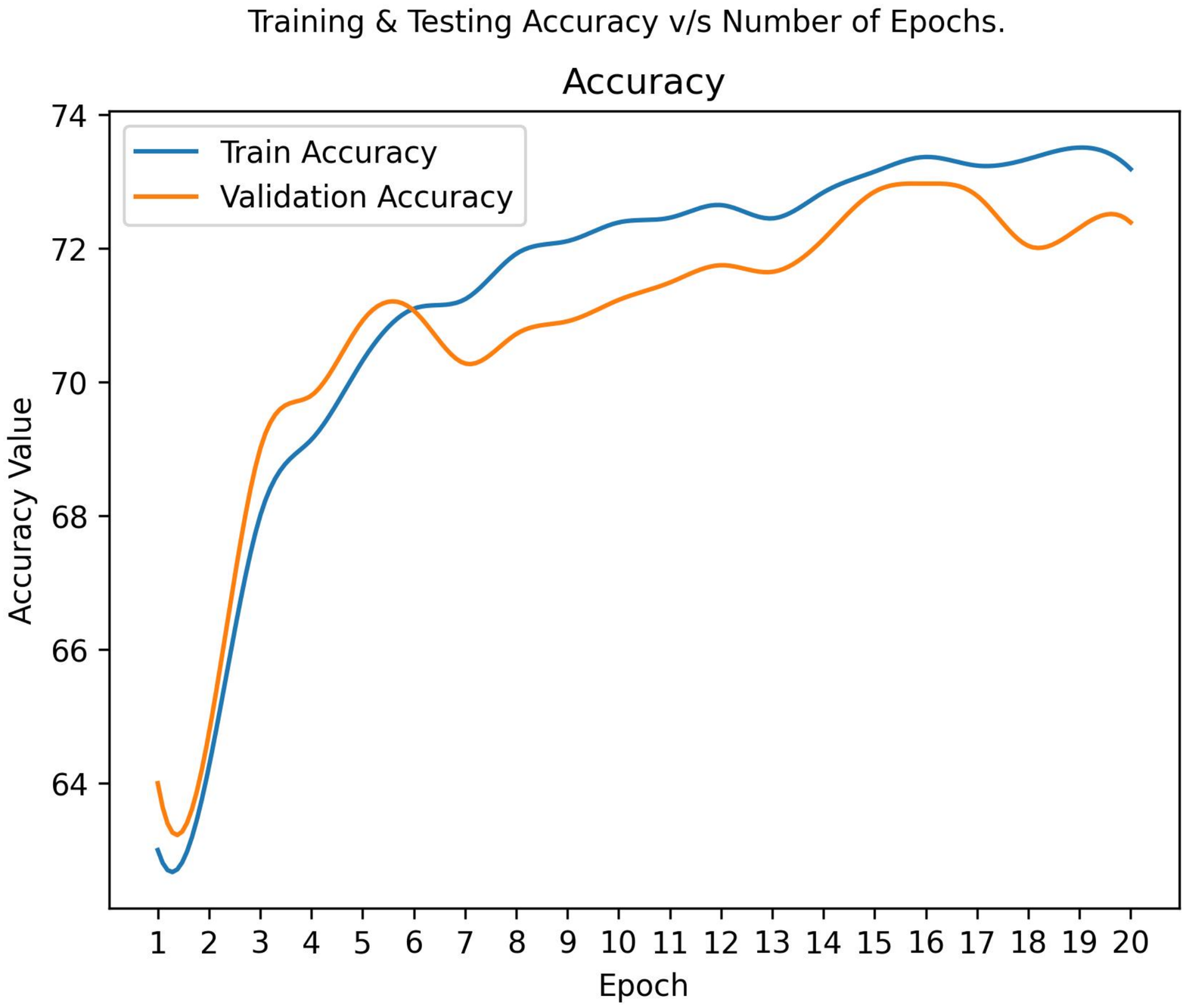} }} \\
\footnotesize Figure 7. The transition graph of training accuracy with increasing number of epochs in MesoNet
\qquad
\subfloat{{\includegraphics[scale=0.25]{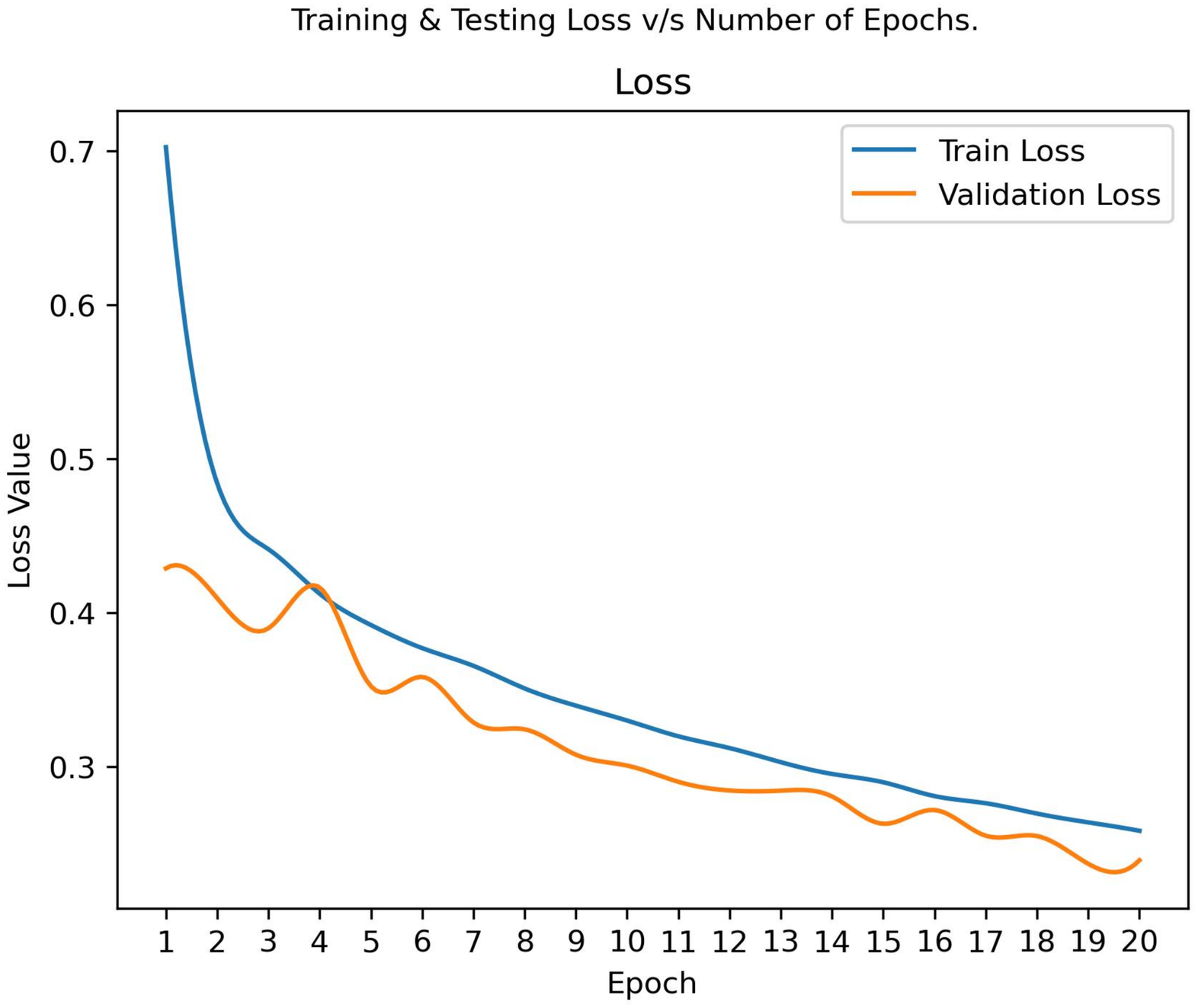} }} \\
\footnotesize Figure 8. The transition graph of training loss with increasing number of epochs in MesoNet
\qquad    
\subfloat{{\includegraphics[scale=0.25]{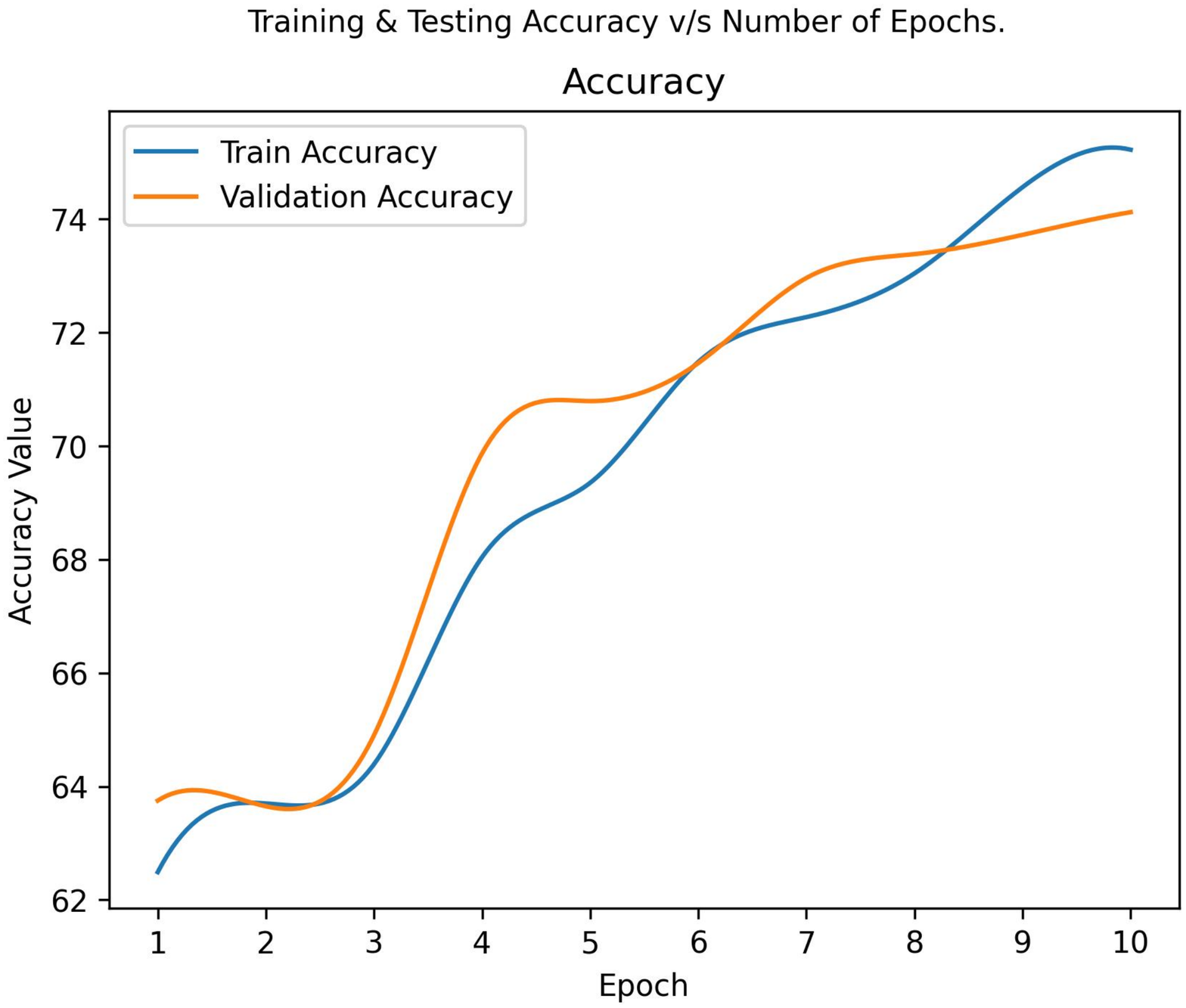} }} \\
\footnotesize Figure 9. The transition graph of training accuracy with increasing number of epochs in ResNet-50
\qquad
\subfloat{{\includegraphics[scale=0.25]{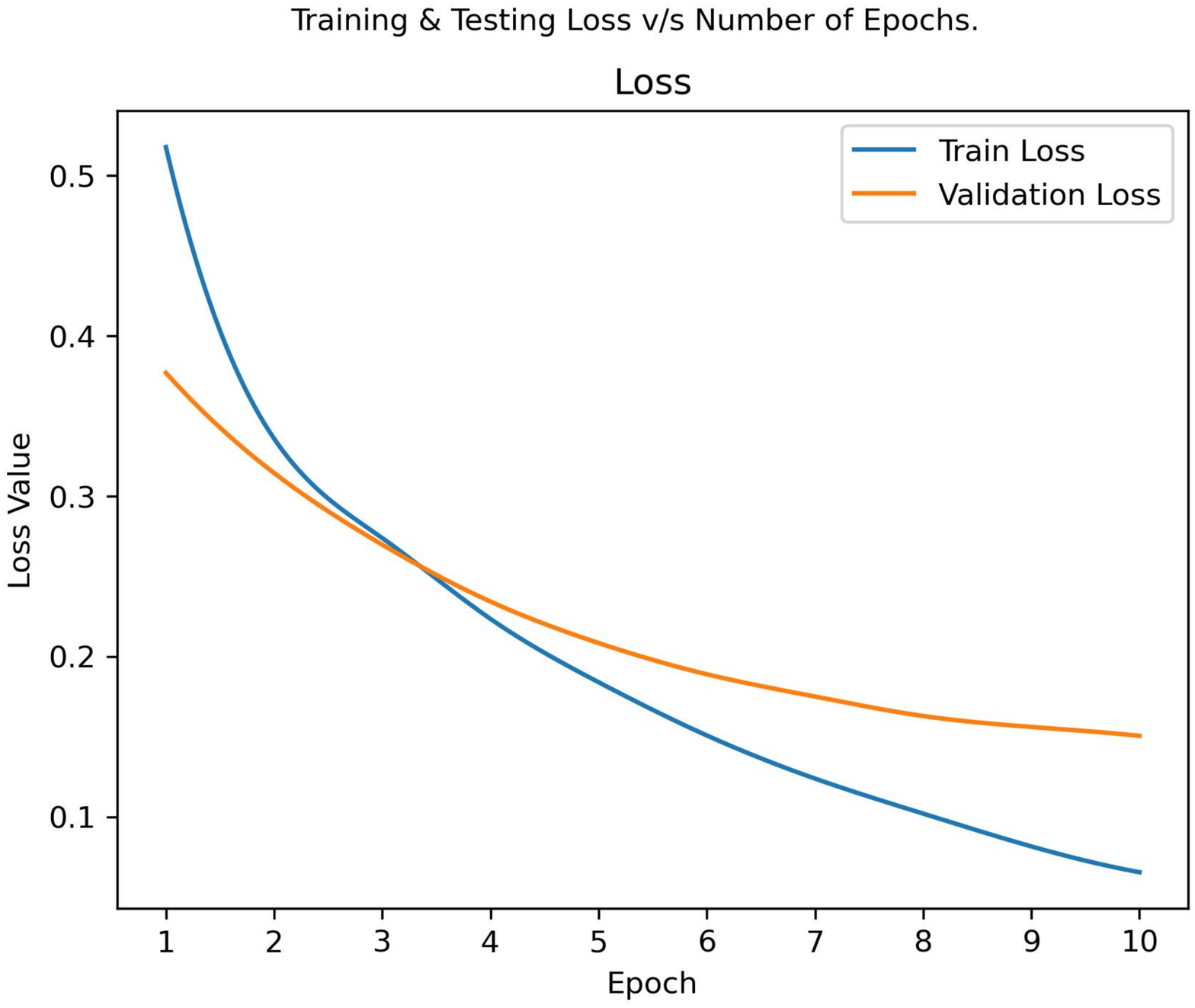} }}  \\  
\footnotesize Figure 10. The transition graph of training loss with increasing number of epochs in ResNet-50
\end{figure}

\begin{figure}
\centering
\subfloat{{\includegraphics[scale=0.25]{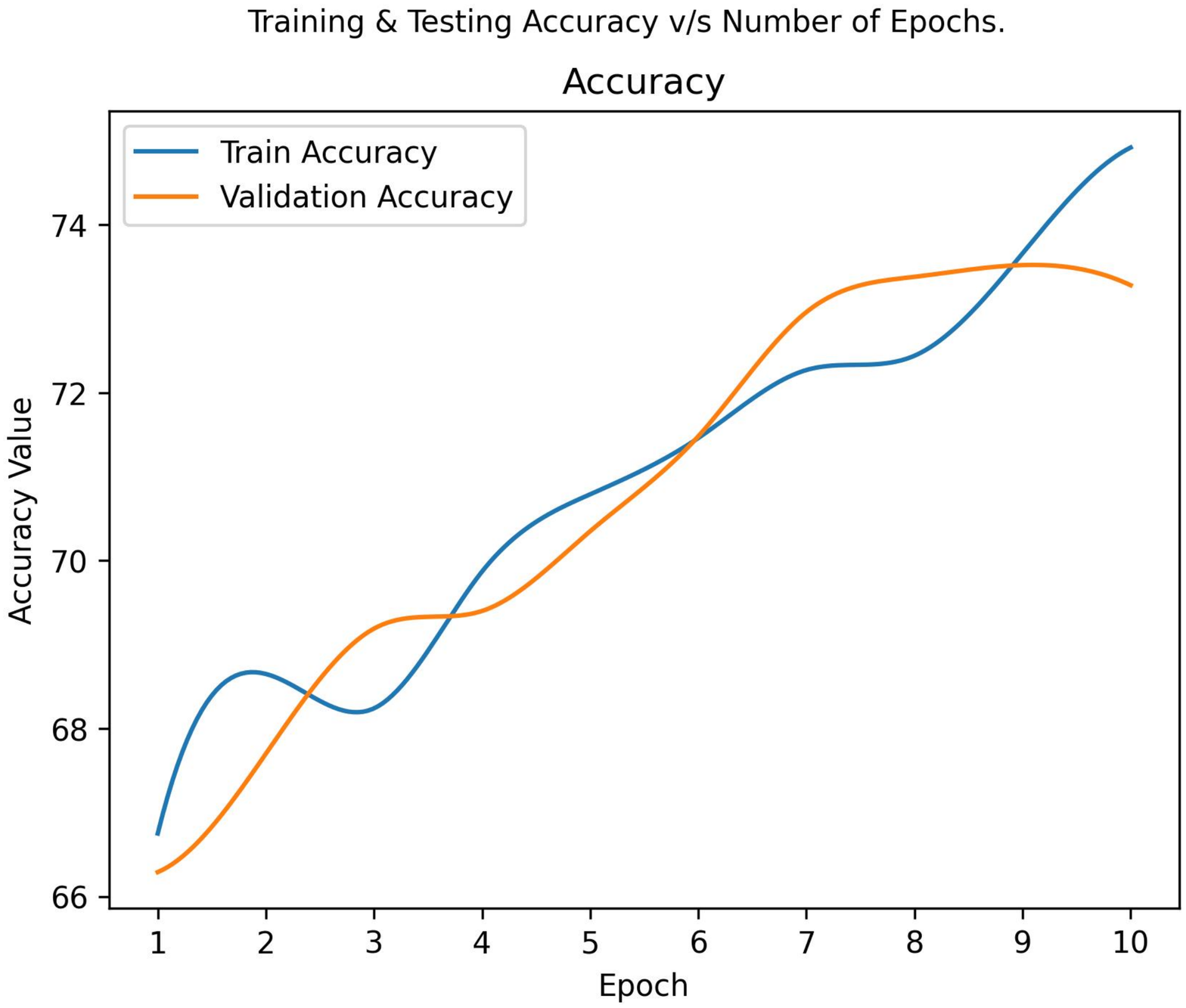} }} \\
\footnotesize Figure 11. The transition graph of training accuracy with increasing number of epochs in VGG-19
\qquad
\subfloat{{\includegraphics[scale=0.25]{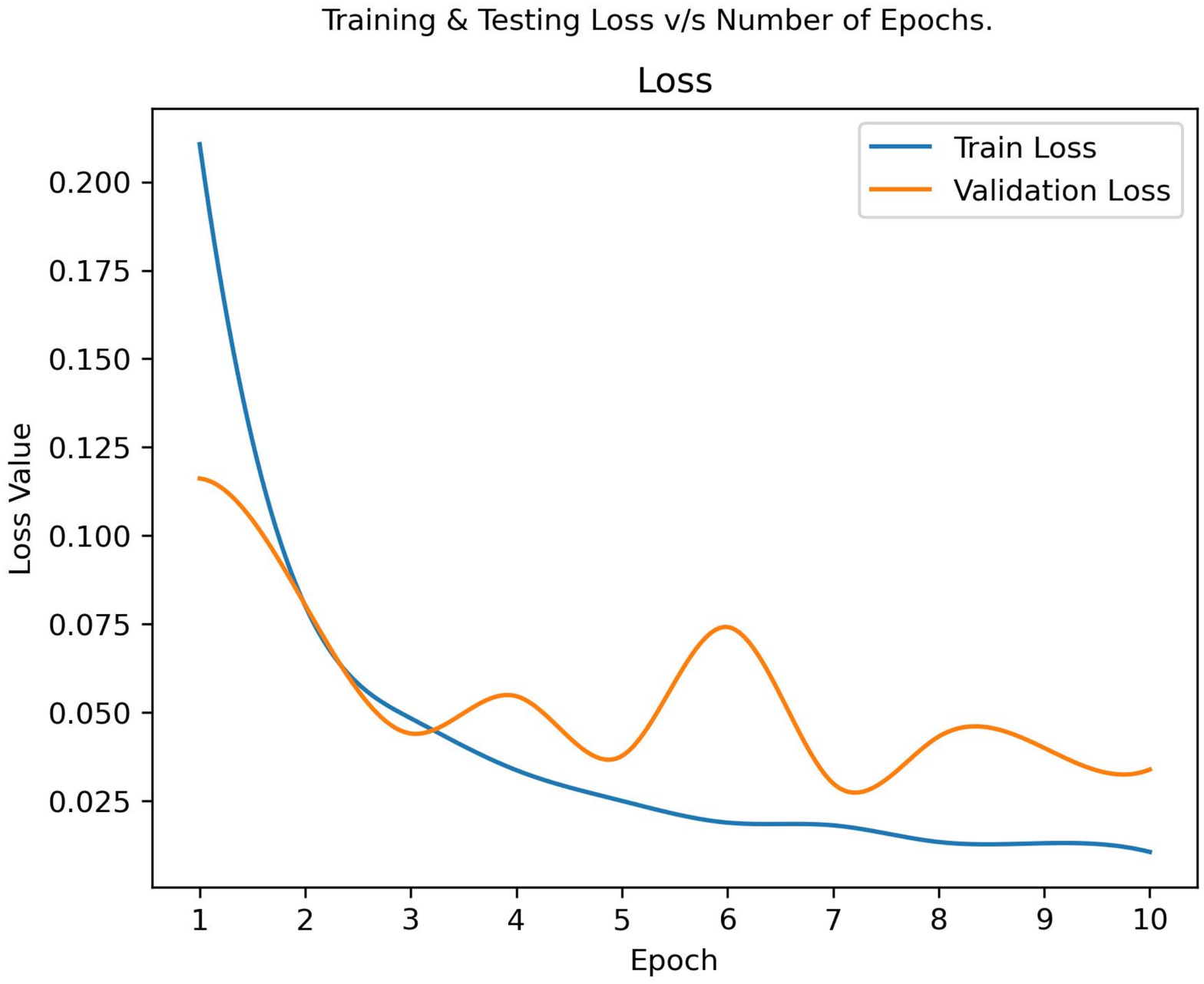} }} \\
\footnotesize Figure 12. The transition graph of training loss with increasing number of epochs in VGG-19
\qquad    
\subfloat{{\includegraphics[scale=0.25]{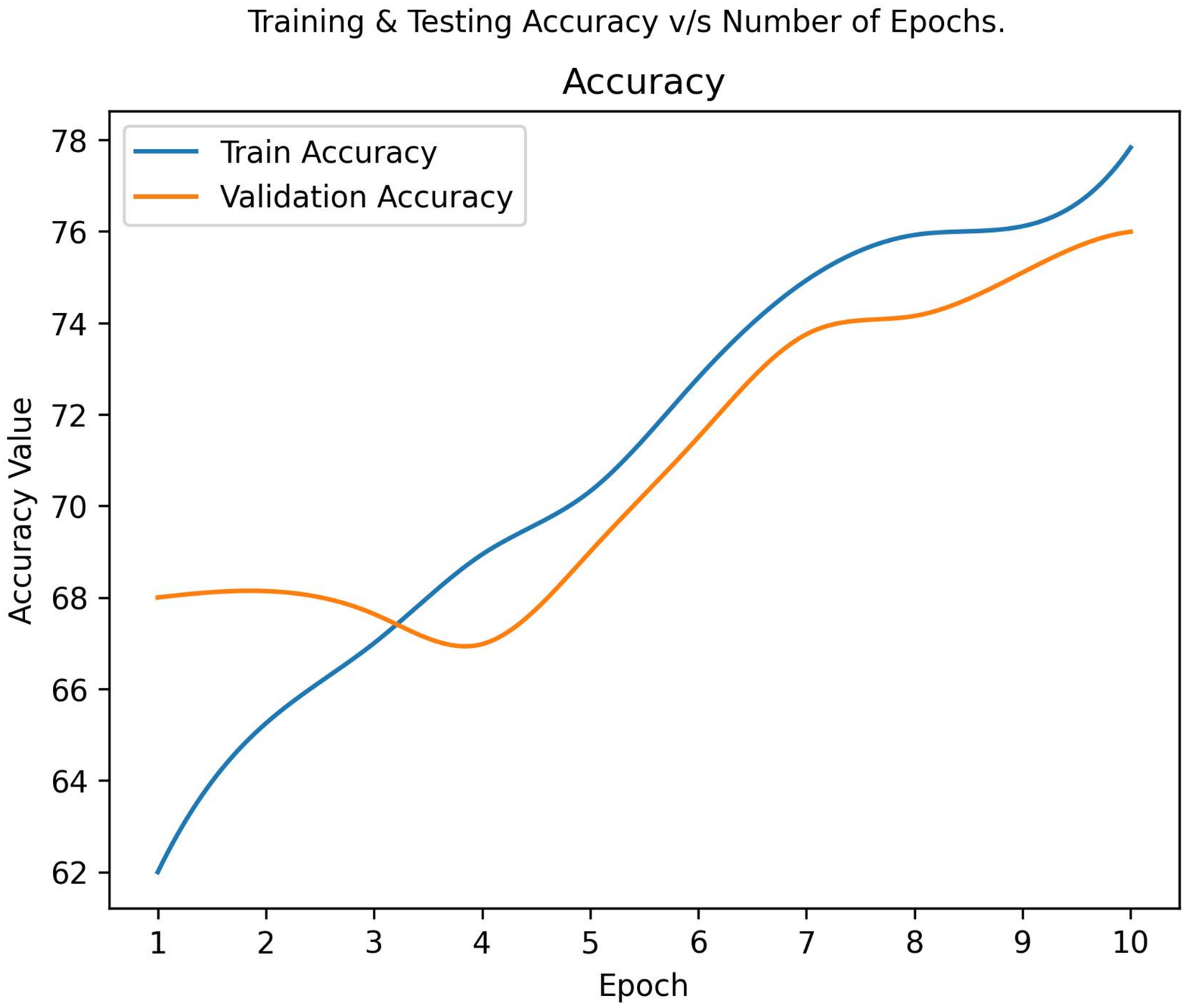} }} \\
\footnotesize Figure 13. The transition graph of training accuracy with increasing number of epochs in Xception
\qquad
\subfloat{{\includegraphics[scale=0.25]{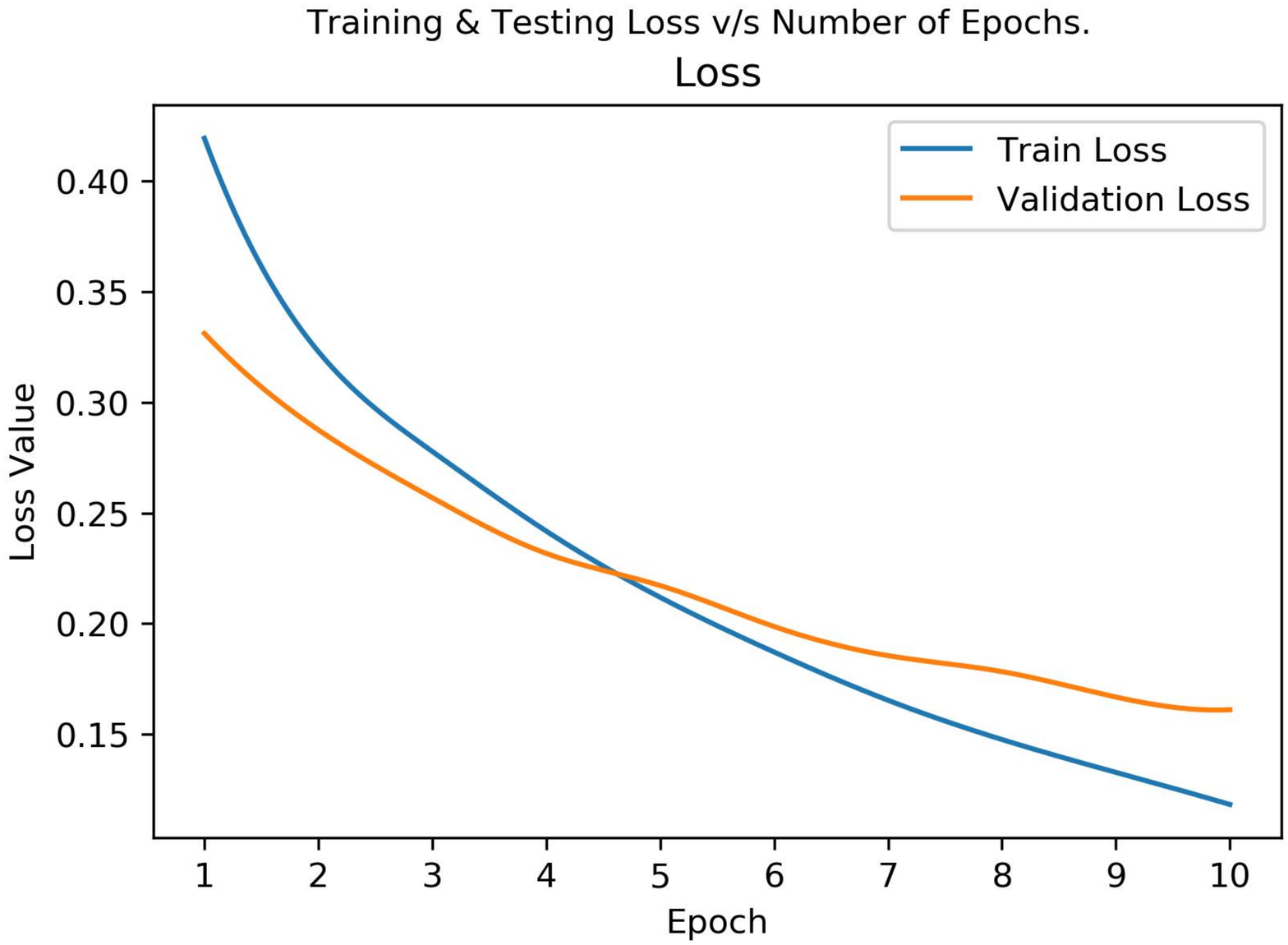} }}  \\  
\footnotesize Figure 14. The transition graph of training loss with increasing number of epochs in Xception
\end{figure}

Furthermore, we visualized the performance of models and how they ameliorated their accuracy and reduced the error rate concerning the number of epochs. All the figures regarding that are found on next page from Figure 7.
\end{small} 
~ \\

\section{\textbf{CONCLUSION}}
~ \\
\begin{small}
In this research, we have surveyed four networks for the task of deep fake detection using a fraction of Celeb-DF, Celeb-DF-v2, and Deep Fake Detection Challenge (DFDC) datasets. We compared them based on their characteristics to appraise the most accurate \& efficient model amongst them. MesoNet is a shallow network \& one of the most basic classifiers used in this group that's why it’s faster than the other networks and in this case, the training accuracy rate it acclaims is also on par with other deeper networks, but due to its simplicity, it’s not possible to classify complex and well-crafted deepfakes as accurately as achieved with other networks. We have found that Resnet-50 \& VGG-19 gave them better results than MesoNet due to their large number of feature extraction layers. When Resnet-50 and VGG-19 are compared with each other, the accuracy rate and loss of both of them lie in the same range but due to more layers in resnet-50, it can perform far better than VGG-19 for Deep Fake detection. At last, the Xception network is unique in its way, because it has a modified depth-wise separable convolution which makes this network flexible \& robust at the same time for this particular problem. That's why, it has delivered better results than other standard networks reviewed in this survey but the drawbacks with the complex networks are that they require more time to train, inference time would be much longer, highly effective dataset, and higher-end hardware is required. Although the inference time can be significantly reduced when used after TensorRT[https://developer.nvidia.com/tensorrt] optimization \\

At this point, Mesonet would only be suggested if the hardware available is low-end, and in the scenario where inference time is more important than the accuracy. However, with the results portrayed by this research, the VGG-19 architecture is most preferable for low to medium end hardware, as not only it provides considerably smaller inference time compared to the other relatively bulky networks(ResNet-50 and Xception) but it is also easier on the hardware, thus making it more viable of a choice when compared to mesonet. But for the scenarios where there are no hardware limitations, the Xception network is the most viable option, as the research concluded, it outperforms the rest of the options considered in this research by a considerable margin, at the same time sacrificing a considerable chunk of time both on training and inference. However, if you find yourself in a very niche scenario where, Xception is coming out to be particularly hard on the resources, while the accuracy VGG-19 provides is not up to standards, then ResNet-50 will prove to be the best option of them all. \\
\end{small}

\section{\textbf{FUTURE ENHANCEMENT}}
\begin{small}
The future development of the applications based on algorithms of deep learning is practically boundless. In the future, we can work on a hybrid algorithm with separate attention-based layers to increase the focus on tampered media than the current set of algorithms with more data to achieve the solutions to the problems.
In the future, the application of these algorithms lies from the public to high-level authorities, as from the differentiation of the algorithms above and with future development, we can attain high-level functioning applications which can be used in the social media companies, classified or government agencies as well as for the common people, we can use these algorithms in different application for ensuring if the media has been tampered with and monitoring the virtual space, The advancement in this field can help us create an environment of safety, awareness, and comfort by using these algorithms in the day-to-day application and high-level application (i.e. Corporate level or Government level). Application-based on artificial intelligence and deep learning is the future of the technological world because of their absolute accuracy and advantages over many major problems. \\
\end{small}

\section{\textbf{ACKNOWLEDGMENT}}
\begin{small}
There are several people without whom this project research work would not have been feasible. Their high academic standards and personal integrity provided me with continuous guidance and support. We owe a debt of sincere gratitude, a deep sense of reverence, and respect to our guide and mentor Prof. Rashid Sheikh, Associate Professor, AITR, Indore for his motivation, sagacious guidance, constant encouragement, vigilant supervision, and valuable critical appreciation throughout this research, which helped us to complete it. We express profound gratitude and heartfelt thanks to Dr. Kamal Kumar Sethi, HOD CSE, AITR Indore for his support, suggestion, and inspiration for carrying out this project. I am very much thankful to other faculty and staff members of the CSE Dept, AITR Indore for providing us all support, help, and advice during this research. We would be failing in our duty if we did not acknowledge the support and guidance received from Dr. S C Sharma, Director, AITR, Indore whenever needed. We are grateful to our parents and family members who have always loved and supported us unconditionally. To all of them, we want to say “Thank you”, for being the best family that one could ever have and without whom none of this would have been possible.
\end{small}

\bibliographystyle{ieeetr}
\bibliography{citation}

\end{document}